\documentclass[runningheads,a4paper]{llncs}

\usepackage{amssymb}
\usepackage{graphicx}
\usepackage{amsmath} 
\usepackage{subcaption}
\usepackage{verbatim}
\usepackage{floatrow}
\newfloatcommand{capbtabbox}{table}[][\FBwidth]

\DeclareCaptionLabelFormat{andtable}{#1~#2  \&  \tablename~\thetable}
\usepackage{lipsum}
\usepackage{etoolbox}

\usepackage{url}
\urldef{\mailsb}\path|{ogil, sanfeliu}@iri.upc.edu|
\newcommand{\keywords}[1]{\par\addvspace\baselineskip
\noindent\keywordname\enspace\ignorespaces#1}

\begin{document}

\mainmatter  

\title{Effects of a Social Force Model reward in Robot Navigation based on Deep Reinforcement Learning }

\titlerunning{Effects of a SFM reward in Robot Navigation based on Deep RL}

\author{\'{O}scar Gil\thanks{Work supported by the Spanish Ministry of Science and Innovation under project ColRobTransp (DPI2016-78957-RAEI/FEDER EU) and by the Spanish State Research Agency through the Mar\'ia de Maeztu Seal of Excellence to IRI (MDM-2016-0656). \'{O}scar Gil is also supported by Spanish Ministry of Science and Innovation under a FPI-grant, BES-2017-082126.}, and Alberto Sanfeliu}

\institute{Institut de Rob\`otica i Inform\`atica Industrial, CSIC-UPC\\
\mailsb}
\authorrunning{}

\maketitle

\vspace{-5mm}

\begin{abstract} 

In this paper is proposed an inclusion of the Social Force Model (SFM) into a concrete Deep Reinforcement Learning (RL) framework for robot navigation. These types of techniques have demonstrated to be useful to deal with different types of environments to achieve a goal. In Deep RL, a description of the world to describe the states and a reward adapted to the environment are crucial elements to get the desire behaviour and achieve a high performance. For this reason, this work adds a dense reward function based on SFM and uses the forces in the states like an additional description. Furthermore, obstacles are added to improve the behaviour of works that only consider moving agents. This SFM inclusion can offer a better description of the obstacles for the navigation. Several simulations have been done to check the effects of these modifications in the average performance.

\keywords{Robot Navigation, Deep Reinforcement Learning, Social Force Model, dense reward function}
\end{abstract}

\vspace{-7mm}
\section{Introduction}\label{sec_introduction}


In Robotics research one of the most important task is improve the ability of a robot to travel to a goal or destination. This ability depends on the challenging that the environment can be, the information about the environment that the robot can obtain and the algorithms which the robot uses to navigate. These factors limit the applicability in most cases.


The situations that can be found in a navigation task are wide and very complex. Different traditional methods have been developed based on potential fields, forces, or other features to solve the task. Actually, one of the approaches to get this purpose is to use Deep Reinforcement Learning techniques.   


These techniques have been used to teach the robot to take the correct actions during navigation but it is difficult to achieve a good performance in complex environments\cite{trautman}. In the particular case of facing complex obstacles SFM can be useful to describe the environment better and avoid them.


In the remainder of the paper, preliminaries and related work is presented in Sections 2 and 3. In Section 4 the problem formulation is exposed. Section 5 explains a new type of reward based on Social Force Model and Section 6 presents the simulations and the results. Finally, conclusions are discussed in Section 7.
\section{Preliminaries}

\subsection{Social Force Model}

For simulate pedestrians motion it is important to consider the physical forces caused by the environment like friction forces to walk, gravity forces or forces in a collision but it is equally important to consider the influence of other pedestrians and obstacles in social rules to walk. For example, when a person who is walking see an obstacle change the trajectory before be near the obstacle to avoid a possible collision. In this process the obstacle does not apply forces to the person but the person reacts as if a force exists.

This type of "virtual forces" that do not exist but can be a way to model the social interactions are known as the Social Force Model\cite{helbing95}. This model simulates the social interactions when a person walks to a determinate goal. There are 2 types of forces in this model applied to an agent $n$(robot or pedestrian). One is the attractive force to the goal, $\textbf{\textit{q}}^{goal}_{n}$,
\begin{equation}
    \textbf{\textit{f}}_{n}(\textbf{\textit{q}}^{goal}_{n})=k\left(\textbf{\textit{v}}_{n}^{0}(\textbf{\textit{q}}^{goal}_{n})-\textbf{\textit{v}}_{n}\right) \ ,
\end{equation}
where $k$ is a constant, $\textbf{\textit{v}}_{n}^{0}(\textbf{\textit{q}}^{goal}_{n})$ is the preferred velocity to the goal and $\textbf{\textit{v}}_{n}$ is the current agent velocity.
The second type of force is the repulsive force of an obstacle or pedestrian $z$ to an agent $n$,
\begin{equation}
    \textbf{\textit{f}}_{z,n}^{int}=a_{z}e^{\left(d_{z}-d_{z,n}\right)/b_{z}}\hat{\textbf{\textit{d}}}_{z,n} \ ,
\end{equation}
where ${a_{z}, b_{z}, d_{z}}$ are parameters of the forces, $d_{z,n}$ is the distance between the agent $n$ and the obstacle or pedestrian $z$ and $\hat{\textbf{\textit{d}}}_{z,n}$ is the unitary vector in the direction defined by $n$ and $z$, pointing to $n$.
The resultant force applied to the agent is:
\begin{equation}
    \textbf{\textit{F}}_{n}= \textbf{\textit{f}}_{n}(\textbf{\textit{q}}^{goal}_{n})+\sum_{z=1}^{Z}\textbf{\textit{f}}_{z,n}^{int} \ .
\end{equation}
Several works have been developed using the SFM in order to get socially acceptable trajectories for robots in diverse situations like side-by-side or approaching tasks\cite{ferrer2014proactive,ferrer2017,repiso2019adaptive} adding elements to incorporate anisotropy or other goals. 

In this work the forces are calculated only for the robot and the resultant force is calculated only with the repulsive forces. The attractive force it's only used in a new type of reward.

\subsection{Optimal Reciprocal Collision Avoidance (ORCA)}

Reciprocal Velocity Obstacle (RVO) and Optimal Reciprocal Collision Avoidance (ORCA) are reactive methods to avoid collisions between moving agents\cite{van2008reciprocal,van2011reciprocal}. In ORCA, the current velocities and positions of all agents in an environment are used to calculate the new velocities that all agents should have to avoid all the possible collisions. 

The advantage of this method is that the ratio of collisions is zero by definition because ORCA chooses the velocities to avoid collisions. 

On the other hand, if an agent is not controlled by ORCA, it cannot be guaranteed that the collision with this agent is prevented. For this reason, ORCA is not very useful to control an agent like a robot that moves with independent agents like people or animals. Additionally, ORCA has problems when an agent has to avoid a concave obstacle. This problem limits the type of obstacles that can be used in the simulations.

In this paper the RVO library with ORCA is used to simulate the environment with agents and obstacles. The robot is controlled with ORCA in the Imitation Learning steps to learn the basic movements to the goal.
\section{Related Work}



In this section, different works related to this paper are briefly described.

\subsection{The reward shaping problem}

It is a very common problem in Reinforcement Learning tasks decide what type of reward choose to improve the training process. This is known as reward design or reward shaping.

Reward shaping is used to decrease the temporal credit assignment problem. This occurs when the agent only receives reward in a goal that must achieve and cannot use this information to know what are the best individual actions to reach the goal. 

Several forms of reward shaping can be used to handle the temporal credit assignment\cite{grzes2011reward,Ng:1999:PIU:645528.657613,metalearning}. It is difficult to get a proper reward that not get stuck in local minima. In this paper it is proposed a different reward transforming a sparse reward into a dense reward based in the SFM.

\subsection{Navigation based on Deep Reinforcement Learning}

Supervised Learning techniques have been applied to predict the movements of agents in a temporal horizon. For example, sequence models that use Long Short Term Memory (LSTM) recurrent neural networks like Social LSTM\cite{alahi2016} and other\cite{sirin2019,hasan2018} are capable to encode the Human-Robot interactions and Human-Human interactions to improve the predictions. Other techniques are based in generative models like Social-GAN\cite{gupta2018} or SoPhie\cite{amir2018} that use pooling modules and attention modules.

This type of techniques offers very useful information in navigation tasks but not take into account all the elements in navigation tasks like obstacles, actions, kinematics or goals.

For cover a more general point of view in the robot navigation task, Reinforcement Learning techniques are a more suitable way to achieve a good performance. Normally, these techniques use neural networks due to the very high number of states to make the computation possible and, for that reason, in these cases are Deep Reinforcement Learning techniques. 

In particular, here it is used the Temporal Difference (TD) algorithm\cite{Sutton1998} using a neural network to calculate the value function. The aim of this method is to use the next prediction to actualize the current prediction before reach the final state taking into account that the states are correlated. This method gives very good results in areas like finance\cite{van2001temporal} and games\cite{tesauro1995temporal}.

Several works\cite{DBLP:journals/corr/ChenELH17,DBLP:journals/corr/ChenLEH16,DBLP:journals/corr/abs-1805-01956,DBLP:journals/corr/abs-1709-10082} for navigation have been developed using these methods to learn policies. Sometimes, these methods combine visual information to improve predictions like PoliNet\cite{hirose2019deep}. Another method\cite{francis2019long} combines Probabilistic Roadmaps and a RL based local planner to guide the robot for long-range indoor navigation.

There are other works that use Imitation Learning approaches to learn policies\cite{liu2018map,long2017deep,tail2018socially}. In the Crowd-Robot Interaction (CRI) approach\cite{chen2018crowd}, in which is mainly based the navigation in this paper, Imitation Learning is combined with the Deep Reinforcement Learning task to obtain a policy (SARL) for robot navigation. For Imitation Learning, the policy used to control all the agents is ORCA. Deep V-learning algorithm is used in the Imitation Learning and Reinforcement Learning phases with a temporal-difference method, a fixed target network and standard experience replay.
In contrast with this work, in the CRI approach obstacles have not been used in the training. It causes some problems in the robot movement when obstacles are added to test the policy.
\section{Problem Formulation}
In this work, it is considered the same navigation task as in the CRI approach. A robot navigates to a goal through an environment with $n$ agents. The difference is that there are obstacles considered too. The objective is to learn a policy to decide the velocities for navigate to a goal avoiding the agents and obstacles with a velocity close to a preferred value and taking socially acceptable trajectories.

The state of the robot is defined with $\textbf{s}$. The states of the agents are represented in $\textbf{w}=[w_{1}, w_{2}... w_{n}]$. In this list of agents, obstacles are included too like agents with velocity equal to zero. All these states are rotated to the robot centred coordinate system with the $x$ axe pointing to the goal. In this coordinate system the states, which proceed from the CRI work, are this,
\begin{equation}
\begin{array}{c}
    \textbf{s}= [d_{g},v_{pref},\theta,r,v_{x},v_{y}] \ , \\
    \textbf{w}_{i}=[p_{x},p_{y},v_{x},v_{y},r_{i},d_{i},r+r_{i}] \ ,
\end{array}
\end{equation}
where $d_{i}=||\textbf{p}-\textbf{p}_{i}||_{2}$ is the robot's distance to the agent $i$, $d_{g}=||\textbf{p}-\textbf{p}_{g}||_{2}$ is the robot's distance to the goal, $\textbf{p}=[p_{x},p_{y}]$ is the agent or obstacle position, $\textbf{v}=[v_{x},v_{y}]$ is the current velocity, $v_{pref}$ is the preferred velocity of the robot, $\theta$ is an angle different to zero for non-holonomic kinematic cases, $r_{i}$ is the agent or obstacle radius and $r$ is the robot radius.

In this work it is used a second type of states that include a SFM description of the agents and obstacles introducing the repulsive forces $\textbf{\textit{f}}_{i}=[f_{x},f_{y}]$ (described in Section 2.1) of each agent and obstacle applied to the robot. These forces are included in the agent states and obstacle states. In the robot state it is included the resultant force of all the agents and obstacles, $\textbf{\textit{F}}=[F_{x},F_{y}]$. These states are,
\begin{equation}
\begin{array}{c}
    \textbf{s}= [d_{g},v_{pref},\theta,r,v_{x},v_{y}, F_{x}, F_{y}] \ , \\
    \textbf{w}_{i}=[p_{x},p_{y},v_{x},v_{y},r_{i},d_{i},r+r_{i}, f_{x},f_{y}] \ .
\end{array}
\end{equation}
The states for all the agents and obstacles are always known and used to calculate the value function using a neural network and the Temporal Difference algorithm of the CRI work\cite{chen2018crowd}. With this value function the greedy policy is calculated, which gives in this case the robot's velocity(the optimal action) each timestep.

\section{Social Force Model Reward}

In this work two types of rewards are used to guide the robot to the correct behaviour. The first reward is the reward used in the CRI approach\cite{chen2018crowd}:
\begin{equation}
    R=\left\{\begin{array}{ll}
        -0.25 & \textrm{if }d_{t}<0\\
        0.25(-0.1+d_{t}/2) & \textrm{else if }d_{t}<0.2\\
        1 & \textrm{else if } \textbf{p}=\textbf{p}_{g}\\
        0 & \textrm{otherwise}
    \end{array}\right. \ ,
\end{equation}
where $d_{t}=d_{i}-r-r_{i}$, represents the distance between agents minus the radius of the 2 agents.

The second reward is a dense reward based on the SFM described in Section 2.1. This Social Force Model Reward takes into account the attractive force to the goal and the repulsive forces of the agents and obstacles,
\begin{equation}
    R=\left\{\begin{array}{ll}
        -0.25 & \textrm{if }d_{t}<0\\
        Ae^{-Bd_{t}} & \textrm{else if }d_{t}<0.2\\
        1 & \textrm{else if } \textbf{p}=\textbf{p}_{g}\\
        k-||k(\textbf{\textit{v}}-\textbf{\textit{v}}_{pref})||_{2}/2-0.0001d_{g} & \textrm{otherwise} 
    \end{array}\right. \ ,
\end{equation}
where $A=-0.03$, $B=10$ and $k=0.001$. These values have been chosen looking for rewards that are not very different from the first reward.
The first term of the reward is applied in collisions, the second term is applied when an agent or obstacle is very near, the third term is applied if the robot get the goal and the last term is equivalent to an attractive force that gives more reward if the velocity is pointing to the goal and the robot is near to the goal. A variation of this reward it is used in a version with a little temporal discount, $-0.02(t-10)$, if $t\geqslant10$ in the third term. This discount is chosen to be not very high and taking into account the average time needed to reach the goal in the best cases.
\section{Experiments}
In this section are described the environment used in the simulations, the metrics to compare different versions, the results and the implementation details. No real experiments have been conducted.
\subsection{Simulation Environment}
The environment in the simulations is created using gym and the RVO library and plotted using matplotlib library like in the CRI work. It contains a robot in an initial position $(0,-4)$ and the robot's goal in $(0,4)$. The agents and the robot are represented like circles using the same radius. 

The obstacles are squares whose side is equal to the diameter of the agents and have an agent inside. This agent inside the obstacle has velocity zero and gives the state's information of the obstacle, $w_{i}$. This is to achieve that the robot be aware of the obstacle when is controlled by the SARL policy. The square obstacle is generated to achieve that the agents controlled by ORCA distinguish between agents and obstacles.  

\begin{figure}[t]
    \begin{center}
    \includegraphics[width=0.45\textwidth]{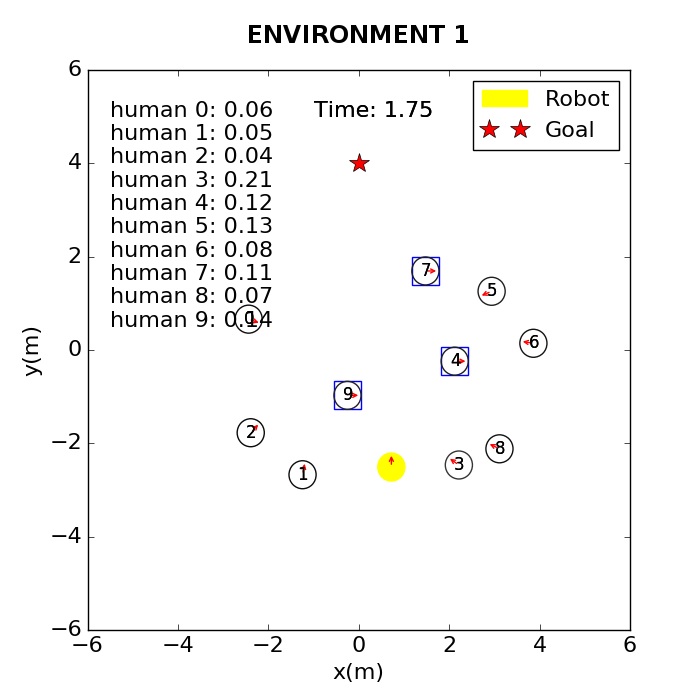}
    \includegraphics[width=0.45\textwidth]{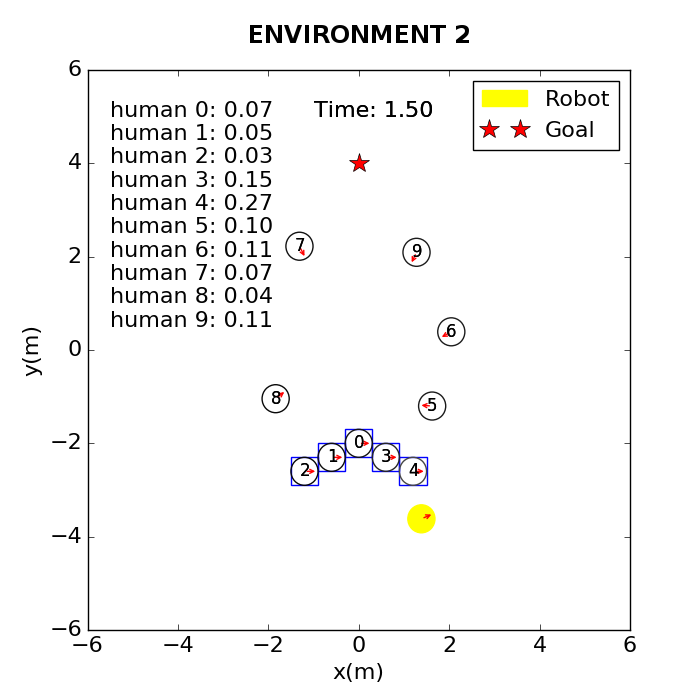}
    \end{center}
    \caption{{\bf First and second environment.} In the first environment obstacles are separated. In the second case, obstacles are joined into a concave barrier.}
    \vspace{-4mm}
\label{nonbarrier}
\end{figure}

\begin{figure}[tb]
    \begin{center}
    \includegraphics[width=0.45\textwidth]{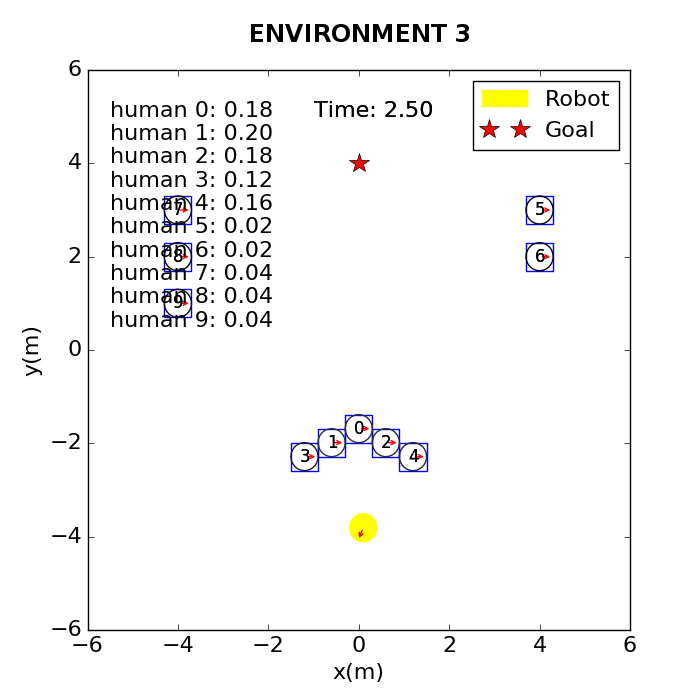}
    \includegraphics[width=0.45\textwidth]{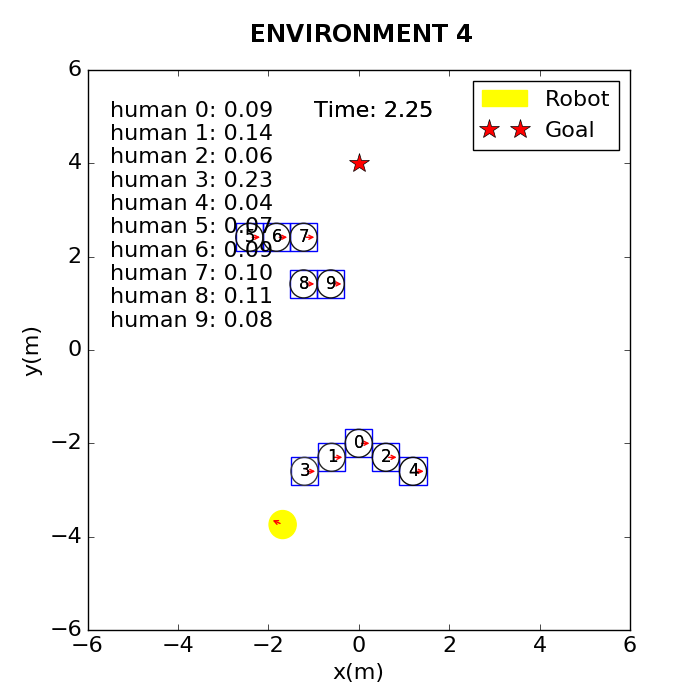}
    \includegraphics[width=0.45\textwidth]{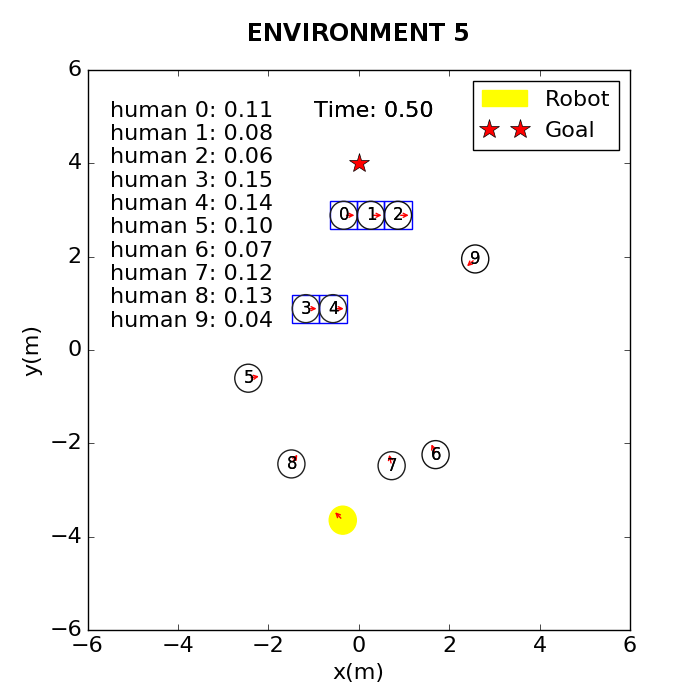}
    \end{center}
    \caption{{\bf Other environments.} Environments from the third to the fifth.}
    \vspace{-4mm}
\label{barrier}
\end{figure}

There are 5 types of environments that always contain 10 elements generated randomly as agents or obstacles. In the first environment, shown in Fig. \ref{nonbarrier}, the agents are generated in a circumference with their goals in the opposite side. The probability of generate an agent is 0.6 and 0.4 for an obstacle.
The second environment, shown in Fig. \ref{nonbarrier}, has always 5 obstacles joined into a fixed concave barrier and 5 agents in a circumference with their goals in the opposite side. The other environments in Fig. \ref{barrier} use combinations of the concave barrier and barriers of 2 or 3 obstacles. In environment 3 the concave barrier can be in different positions between the robot and the goal. 
\subsection{Metrics}
In this section are explained the metrics used to measure the performance of the navigation. These metrics are the same than in the CRI work.
\begin{itemize}
    \item \textbf{Success Rate}: This value measures the ratio of times that the robot reaches the goal.
    \item \textbf{Collision Rate}: This value measures the ratio of times that the robot collides with agents or obstacles.
    \item \textbf{Navigation Time}: Time used for the robot to reach the goal. The maximum time is 25 s.
    \item \textbf{Total Reward}: The average cumulative reward of all the episodes.
\end{itemize}
The most meaningful metrics are the success rate and collision rate because brief on the most important features of the navigation task.
\subsection{Quantitative Evaluation}
\begin{figure}[t]
    \begin{center}
    \includegraphics[width=0.8\textwidth]{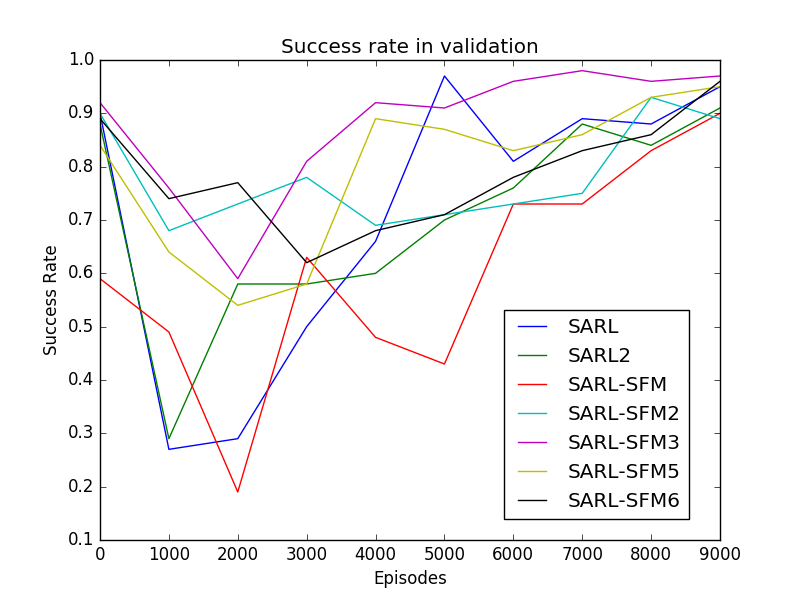}
    \end{center}
    \caption{{\bf Validation graphic.} This graphic shows the validation results for the success rate each 1000 episodes of the first type of training.}
    \vspace{-4mm}
\label{val}
\end{figure}

In Table \ref{table1} the average results of a test using 500 episodes are exposed for the different versions:

\begin{itemize}
    \item \textbf{SARL}: The CRI approach without local map. 
    \item \textbf{SARL-SFM}: With SFM reward and temporal discount in the reward.
    \item \textbf{SARL-SFM2}: With SFM reward.
    \item \textbf{SARL-SFM3}: With SFM reward but without concave barrier in training.
    \item \textbf{SARL-SFM4}: With SFM reward but $k=0.003$ in the reward.
    \item \textbf{SARL-SFM5}: With forces included in the states and without SFM reward.
    \item \textbf{SARL-SFM6}: With forces included in the states and with the SFM reward.
\end{itemize}
\begin{table}[tb]
\begin{center}
\begin{tabular}{|c|c|c|c|c|}
\hline
\ Version \ & \ Success Rate \ & \ Collision Rate \ & \ Navigation Time \ & \ Total Reward \ \\
\hline \hline
\textbf{SARL(2 times)} & \textbf{0.96/0.97} & 0.02/0.01 & \textbf{10.85}/11.77 & \textbf{0.3059/0.2850}\\ \hline
\textbf{SARL-SFM} & 0.91 & 0.03 & 11.83 & 0.2417\\ \hline
\textbf{SARL-SFM2} & 0.86 & 0.02 & 11.92 & 0.2363 \\ \hline
\textbf{SARL-SFM3} & 0.93 & 0.01 & 11.27 & 0.2741 \\ \hline
\textbf{SARL-SFM4} & 0.90 & \textbf{0.00} & 11.61 & 0.2646\\ \hline
\textbf{SARL-SFM5} & \textbf{0.96} & 0.03 & 11.69 & 0.2799\\ \hline
\textbf{SARL-SFM6} & \textbf{0.96} & 0.03 & 11.25 & 0.2713\\ \hline
\end{tabular}
\caption{{\bf Test results.} Average results obtained in the test evaluation after 500 episodes. All these cases are obtained using only the two first environments.}
\label{table1}
\end{center}
\end{table}
\begin{table}[tb]
\begin{center}
\begin{tabular}{|c|c|c|c|c|}
\hline
\ Version \ & \ Success Rate \ & \ Collision Rate \ & \ Navigation Time \ & \ Total Reward \ \\
\hline \hline
\textbf{SARL} & \textbf{0.901} & \textbf{0.001} & \textbf{10.54} & \textbf{0.3046}\\ \hline
\textbf{SARL-SFM6} & 0.893 & 0.004 & 10.88 & 0.2763\\ \hline
\end{tabular}
\caption{{\bf Second test results.} Average results obtained in the test evaluation after 1000 episodes using all the environments in training.}
\label{table2}
\end{center}
\end{table}
In the Fig. \ref{val} is shown the success rate for each type of training during the validation. In Table \ref{table2} are shown the results of a second type of training with only the SARL and SARL-SFM6 cases using all the environments.

\subsection{Qualitative Evaluation}

Comparing to the cases analysed in the CRI paper, in which there were only 5 moving agents without obstacles, this paper has proposed a more demanding environment with 10 elements between agents and obstacles. Normally, with the new environments in training, the robot does not get stuck because of the obstacles and can surpass the concave barrier.

In spite of the fact that the robot faces better the obstacles, there are some cases where the robot is very slow avoiding the concave barrier. If there are not agents moving in the environment the robot gets stuck with the concave barrier.

In Table \ref{table1}, results obtained are, in general, worse than in the CRI paper because of the more challenging environment and the best success rate is obtained without the SFM reward. Comparing the SARL-SFM2 to SARL-SFM3, although the success rate is better in SARL-SFM3, in SARL-SFM2 the robot get stuck fewer in the concave barrier episodes. For this reason, the concave barrier can be useful in the training phase. 

In the SARL-SFM4, k is 3 times bigger to increment the rewards and the influence of the attractive effect to the goal and there is not temporal discount. This causes a worse performance than the SARL-SFM case.  

Although the test results are better in the SARL case, the training progression is not equal for all the cases. This can be observed in Fig. \ref{val}. For example, in the SARL-SFM3 the validation applied after 7000 episodes of training gives a success rate of 0.98. In SARL-SFM5, this success rate after 4000 episodes is 0.89, while in the SARL case was only 0.66. In the SARL-SFM case the success rate is always smaller than the SARL success rate.

The 2 last cases, SARL-SFM5 and SARL-SFM6, combine the SFM reward with the new states and offers better results that cases with only SFM reward. The convergence is faster in the first part of the training than the SARL case and the performance in test is more or less the same. These results suggest that the training progression is faster introducing the forces in the states and the oscillations in training are reduced.

In Table \ref{table2} all the environments are used in the training and the performance of the robot facing obstacles improves only a little in the SARL and SARL-SFM6 cases. There are more or less a $10 \ \%$ of cases where the robot get stuck and these cases normally occur in environments that only contain obstacles.

\subsection{Implementation Details}
For the implementation, the robot is visible for the agents and the different versions that are compared do not use the local map of the CRI work. The neural network is the same as the network used in the CRI approach. It is only changed the input layer in the first multilayer perceptron (MLP) and the input layer in the last MLP to use the new states proposed in this paper. In the states, the parameters of the repulsive forces described in Section 2.1 are $a_{z}=1$, $d_{z}=0$ and $b_{z}=1$. This choice offers inputs neither too big nor too small to the neural network. 

In the optimization task it is used the Stochastic Gradient Descent method with momentum 0.9. The batch size is 100.

For imitation learning, 3000 episodes are collected using ORCA for the robot and the policy is trained with 50 epochs and learning rate 0.01. In this phase, the environment is the one without the concave barrier in all cases.

For reinforcement learning, the learning rate is 0.001, the discount factor is 0.9, the number of episodes is 10.000 and the policy used is the $\epsilon$-greedy policy of the CRI work. Each 1000 episodes there is a validation phase with 100 episodes. At the end of the training there is a test phase with 500 episodes. In Table \ref{table1}, the environment used for the reinforcement learning phase of the training, the validation phase and the test phase is randomly chosen each episode between one of the two first environments with probabilities 0.7 for the first environment and 0.3 for the second one. In Table \ref{table2}, the environments used in reinforcement learning are a total of 4, from the second to the fifth, with probabilities 0.25, the number of training episodes is 15.000 and there are 1000 test episodes.

The work assumes holonomic kinematics and the same action space that is used in CRI work. The implementation has been developed in PyTorch and the simulations have been launched in a Tesla K40c GPU.

\section{Conclusion}

In this work, 3 modifications have been done in order to check the changes in the performance of the SARL policy, designed in the CRI approach.

Firstly, the environment has been expanded through obstacles to improve the robot's behaviour when faces them. This type of modification has slightly improved the robot behaviour but there are still some problems when the robot faces only obstacles and a better description of the environment or a different neural network would be required.

Secondly, the reward signal has been modified taking account the SFM. This type of modification has caused overall worse results in the navigation performance reducing the success rate and increasing the collision rate. In spite of this problem, the convergence improves in the first 5000 episodes. This better initial convergence could be caused because the algorithm has additional information in the first part of the training.

Finally, the last modification has been the inclusion of the repulsive forces in the agent states and the resultant force in the robot state. This modification has not produced notable changes but it appears that the convergence is faster in the first 5000 episodes. It also makes the test results very much independent of using SFM reward, palliating its negative effect.

To sum up, this approach offers a way to slightly accelerate the convergence in the TD learning for navigation tasks. For better results in the performance of the policy are suggested as future work, forms of reward shaping, other RL algorithms like Policy Gradients, more prediction information of the environment like in Social GAN work\cite{gupta2018} and mix the RL algorithms with a global planner\cite{francis2019long}.

\bibliographystyle{splncs03}
\bibliography{collab.bib}

\end{document}